\documentclass[11pt,a4paper]{article}
\usepackage[newfloat,frozencache,cachedir=.]{minted}

\usepackage{amsmath,amsfonts,bm}

\def\eqref#1{equation~\ref{#1}}

\def\1{\bm{1}}

\def\mD{{\bm{D}}}

\def\mT{{\bm{T}}}

\def\mX{{\bm{X}}}
\def\mY{{\bm{Y}}}

\DeclareMathAlphabet{\mathsfit}{\encodingdefault}{\sfdefault}{m}{sl}
\SetMathAlphabet{\mathsfit}{bold}{\encodingdefault}{\sfdefault}{bx}{n}

\usepackage[hyperref]{emnlp2020}
\usepackage{times}
\usepackage{latexsym}
\usepackage{graphicx}
\usepackage{float}
\usepackage{listings}
\usepackage[nameinlink,capitalize]{cleveref}
\usepackage[inline]{enumitem}
\usepackage{algorithm,algorithmicx}
\usepackage[noend]{algpseudocode}
\usepackage{url}
\usepackage{booktabs}
\usepackage{amsmath}
\usepackage{booktabs}
\usepackage{titling}
\usepackage[colorinlistoftodos,prependcaption,textsize=tiny]{todonotes}

\usepackage{siunitx}
\ExplSyntaxOn
  \cs_new_eq:NN \calc \fp_eval:n
\ExplSyntaxOff

\usepackage{microtype}

\newenvironment{code}{\captionsetup{type=listing,font=small,skip=0pt}}{}

\aclfinalcopy 

\newcommand{\simuleval}{\textsc{SimulEval} }
\newcommand{\simulevalnsp}{\textsc{SimulEval}}

\title{\simuleval: An Evaluation Toolkit \\for Simultaneous Translation}

\author{
  Xutai Ma $^{1, 2}$, Mohammad Javad Dousti$^1$, Changhan Wang$^1$, Jiatao Gu$^1$, Juan Pino$^1$ \vspace*{0.2cm} \\
$^1$Facebook AI\\
$^2$Johns Hopkins University \vspace*{0.2cm} \\
  {\tt xutai\_\thinspace ma@jhu.edu} \\
  {\tt \{juancarabina,dousti,changhan,jgu\}@fb.com} }

\date{}

\begin{document}
\maketitle
\begin{abstract}
Simultaneous translation on both text and speech focuses on a real-time and low-latency scenario
where the model starts translating before reading the complete source input.
Evaluating simultaneous translation models is more complex than offline models
because the latency is another factor to consider in addition to translation quality. The research community, despite its growing focus on novel modeling approaches to simultaneous translation, currently lacks a universal evaluation procedure. Therefore, we present \simulevalnsp, an easy-to-use and general evaluation toolkit for both simultaneous text and speech translation.
A server-client scheme is introduced to create a simultaneous translation scenario, where the server sends source input and receives predictions for evaluation and the client executes customized policies.
Given a policy, it automatically performs simultaneous decoding
and collectively reports several popular latency metrics.
We also adapt latency metrics from text simultaneous translation to the speech task.
Additionally, \simuleval is equipped with a visualization interface to provide better understanding of the simultaneous decoding process of a system. \simuleval has already been extensively used for the IWSLT 2020 shared task on simultaneous speech translation. Code will be released upon publication.
\end{abstract}
\section{Introduction}
Simultaneous translation,
the task of generating translations before reading the entire text or speech source input, has become an increasingly popular topic for both text and speech translation~\cite{grissom2014don, cho2016can, gu2017learning, alinejad2018prediction, arivazhagan-etal-2019-monotonic, zheng-etal-2019-simultaneous,ma2020monotonic, ren-etal-2020-simulspeech}.
Simultaneous models are typically evaluated from quality and latency perspective.
Note that the term \textit{latency} is overloaded and sometimes refers to the actual system speed. In this paper, \textit{latency} refers to the simultaneous ability, which is how much partial source information is needed to start the translation process.

While the translation quality is usually measured by BLEU~\cite{papineni2002bleu, post-2018-call},
a wide variety of latency measurements have been introduced,
such as
Average Proportion (AP)~\cite{cho2016can}, Continues Wait Length (CW)~\cite{gu2017learning}, Average Lagging (AL)~\cite{ma-etal-2019-stacl}, Differentiable Average Lagging (DAL)~\cite{cherry2019thinking}, and so on.
Unfortunately, the latency evaluation processes across different works are not consistent: 1) the latency metric definitions are not precise enough with respect to text segmentation; 2) the definitions are also not precise enough with respect to the speech segmentation, for example some models are evaluated on speech segments~\cite{ren-etal-2020-simulspeech} while others are evaluated on time duration~\cite{ansari-etal-2020-findings}; 3) little prior work has released implementations of the decoding process and latency measurement.
The lack of clarity and consistency of the latency evaluation process
makes it challenging to compare different works and prevents tracking the scientific progress of this field.

In order to provide researchers in the community with a standard, open and easy-to-use method to evaluate simultaneous speech and text translation systems, we introduce \simulevalnsp, an open source evaluation toolkit which automatically simulates a real-time scenario
and evaluates both latency and translation quality. 
The design of this toolkit follows a server-client scheme, which has the advantage of creating a fully simultaneous translation environment and is suitable for shared tasks such as the IWSLT~2020 shared task on simultaneous speech translation\footnote{\url{http://iwslt2020.ira.uka.de/doku.php?id=simultaneous_translation}} or the 1st Workshop on Automatic Simultaneous Translation at ACL 2020\footnote{\url{https://autosimtrans.github.io/}}.
The server provides source input (text or audio) upon request from the client, receives predictions from the client and returns different evaluation metrics when the translation process is complete. The client contains two components, an agent and a state,
where the former executes the system's policy and the latter keeps track of information necessary to execute the policy as well as generating a translation.
\simuleval has built-in support for quality metrics such as BLEU~\cite{papineni2002bleu,post-2018-call}, TER~\cite{snover2006study} and METEOR~\cite{banerjee2005meteor}, 
and latency metrics such as AP, AL and DAL. 
It also support customized evaluation functions.
While all latency metrics have been defined for text translation, we discuss issues and solutions when adapting them to the task of simultaneous speech translation. 
Additionally, \simuleval users can define their own customized metrics. 
\simuleval also provides an interface to visualize the policy of the agent. 
An interactive visualization interface is implemented to illustrate the simultaneous decoding process.
The initial version of \simuleval was used to evaluate submissions from the first 
shared task on simultaneous speech translation at IWSLT 2020~\cite{ansari-etal-2020-findings}.


In the remainder of the paper, we first formally define the task of simultaneous translation. Next, latency metrics and their adaptation to the speech task are introduced. After that, we provide a high-level overview of the client-server design of \simulevalnsp. Finally, usage instructions and a case study are provided before concluding.
\section{Task Formalization}
An evaluation corpus for a translation task contains one or several instances, each of which consists of a source sequence $\mX = [x_1,...,x_{|\mX|}]$ and a reference sequence $\mY^* = [y^*_1,...,y^*_{|\mY|}]$.
The system to be evaluated takes $\mX$ as input, and generates $\mY=[y_1,...,y_{|\mY|}]$.
We denote the elements of the $\mX$, $\mY$ and $\mY^*$  segments.
For text translation, each $x_j$ is an individual word while for speech translation, $x_j$ is a raw audio segment of duration $T_j$.
In the simultaneous translation task, a system starts generating a hypothesis with partial input only.
Then it either reads a new source segment, 
or writes a new target segment.
Assuming $\mX_{1:j} = [x_1,...,x_j], j< |\mX|$ has been read when generating $y_i$,
we define the delay of $y_i$ as
\begin{equation}
    d_i = 
    \begin{cases}
    j, &\text{when input is text}  \\
    \sum_{k=1}^j T_k, &\text{when input is speech} \\
  \end{cases}
\end{equation}
Similar to an offline model, the quality is measured by comparing the hypothesis $\mY$ to the reference $\mY^*$ after the translation process is complete. 
On the other hand, the latency measurement involves considering partial hypotheses.
The latency metrics are calculated from a function which takes a sequence of delays $\mD = [d_1, ..., d_{|\mY|}]$ as input.

\section{Latency Evaluation}
\subsection{Existing Text Latency Metrics}
\label{sec:text_metrics}
First, we review three latency metrics previously introduced for the text translation task.
\begin{description}[style=unboxed,leftmargin=0cm]
\item[Average Proportion (AP)]~\cite{cho2016can}, defined in \cref{eq:ap}, measures the average of proportion of source input read when generating a target prediction.
\begin{equation}
    \text{AP} = \frac{1}{|\mX||\mY|} \sum_{i=1}^{|\mY|} d_i
    \label{eq:ap}
\end{equation}
Despite AP's simplicity, 
several concerns have been raised. Specifically, AP is not length invariant, i.e.\ the value of the metric depends on the input and output lengths. For instance, AP for a wait-$3$ model~\cite{ma-etal-2019-stacl} is 0.72 when $|\mX|=|\mY|=10$ but 0.52 when $|\mX|=|\mY|=100$. Moreover, AP is not evenly distributed on the $[0,1]$ interval, i.e., values below 0.5 represent models that have lower latency than an ideal policy, and an improvement of 0.1 from 0.7 to 0.6 is much more difficult to obtain than the same absolute improvement from 0.9 to 0.8~\cite{ma-etal-2019-stacl}.

\item[Average Lagging (AL)] first defines an ideal policy,
which is equivalent to a wait-0 policy that has the same prediction as the system to be evaluated.
\citet{ma-etal-2019-stacl} define AL as
\begin{equation}
    \text{AL} = \frac{1}{\tau(|\mX|)} \sum_{i=1}^{\tau(|\mX|)} d_i - \frac{\left(i-1\right)}{\gamma}
    \label{eq:al}
\end{equation}
where $\tau(|\mX|) = \text{min}\{i | d_i=|\mX|\}$ is the index of the target token when the policy first reaches the end of the source sentence
and $\gamma = |\mY| / |\mX| $.
$\left(i-1\right) / \gamma$ term is the ideal policy for the system to compare with. AL has good properties such as being length-invariant and intuitive. 
Its value directly describes the lagging behind the ideal policy.

\item[Differentiable Average Lagging (DAL)] introduces a minimum delay of $1 / \gamma$ after each operation.
Unlike AL, it considers the tokens when $i > \tau(|\mX|)$ \cite{cherry2019thinking}.
It is defined in \cref{eq:dal}:
\begin{equation}
    \text{DAL} = \frac{1}{|\mY|} \sum_{i=1}^{|\mY|} d'_i - \frac{i - 1}{\gamma},
    \label{eq:dal}
\end{equation}
where
\begin{equation}
    d'_i =
    \begin{cases}
        d_i & i=0\\
        \max(d_i, d'_{i-1} + \gamma) & i > 0
    \end{cases}.
    \label{eq:dal-2}
\end{equation}
A minimum delay prevent DAL recovering from lagging once it has been incurred.
\end{description}

\subsection{Adapting Metrics to the Speech Task}
In this section, we adapt the three latency metrics introduced in \cref{sec:text_metrics} to the simultaneous speech translation task.
\begin{description}[style=unboxed,leftmargin=0cm]
\item[Average Proportion] is straightforward to adapt to the speech task and as follows:
\begin{equation}
    \text{AP}_{\text{speech}} = \frac{1}{|\mY|\sum_{j=1}^{|\mX|}T_j}  \sum_{i=1}^{|\mY|} d_i
    \label{eq:ap_speech}
\end{equation}

\item[Average Lagging] is adapted as follows:
\begin{equation}
    \text{AL}_{\text{speech}} = \frac{1}{\tau'(|\mX|)} \sum_{i=1}^{\tau'(|\mX|)} d_i - d^*_i,
    \label{eq:al_speech}
\end{equation}
where $\tau'(|\mX|) = \text{min}\{i | d_i=\sum_{j=1}^{|\mX|}T_j\}$ and $d^*_i$ are the delays of an ideal policy, of which the straightforward adaption is  
$d^*_i = (i-1)\times\sum_{j=1}^{|\mX|}T_j\ / |\mY|$.
However such adaptation is not robust for models that tend to stop hypothesis generation too early and generate translations that are too short.
This is more likely to happen in simultaneous speech translation where a model can generate the end of sentence token too early, for example when there is a long pause even though the entire source input has not been consumed.
\begin{figure}
    \centering
    \includegraphics[width=0.5\textwidth]{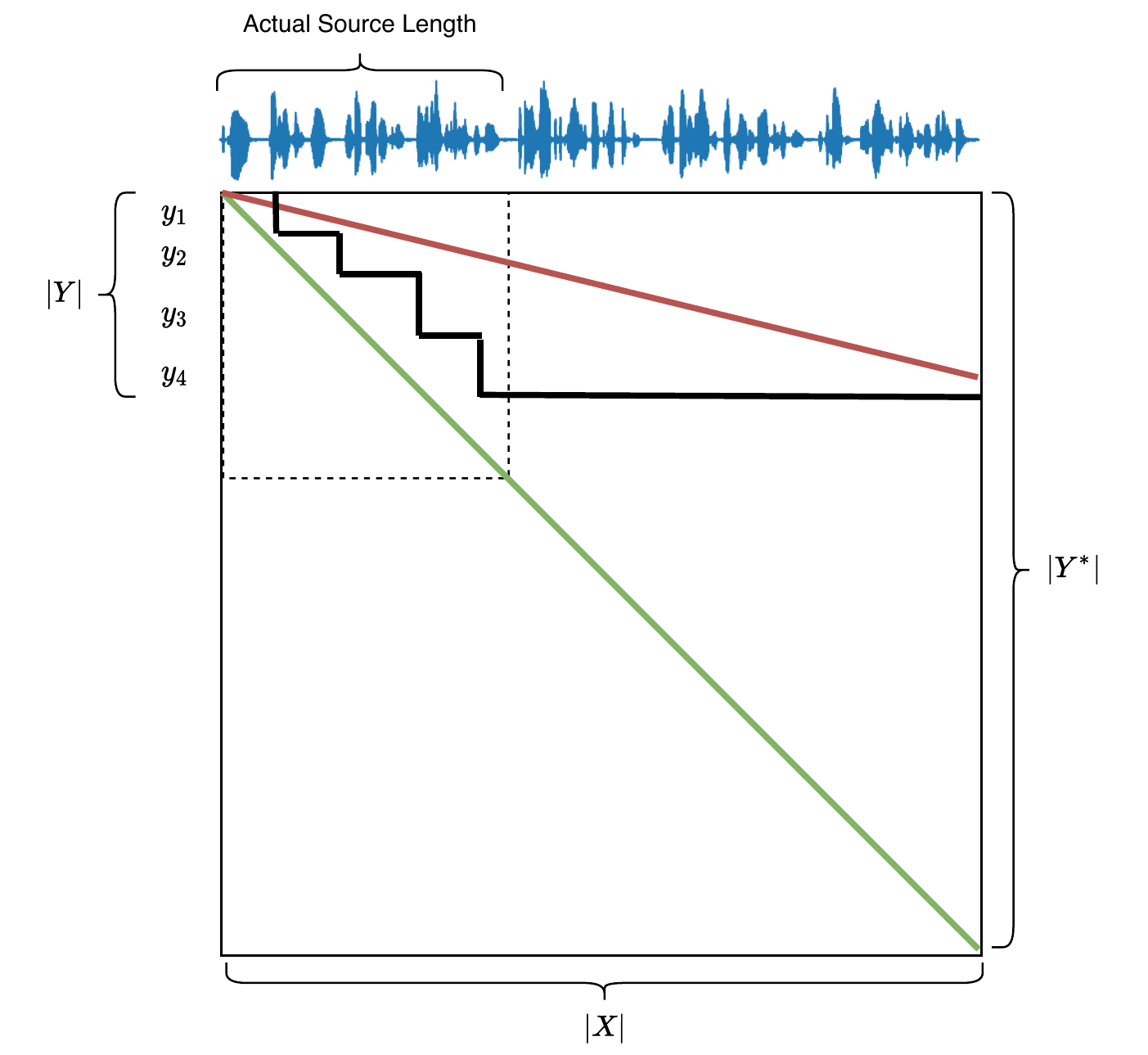}
    \caption{
        An example of original AL failed on early stop translation.
        Red  line shows the ideal policy in \cite{ma-etal-2019-stacl}.
        Green line depicts the modified ideal policy in this paper.
        Black demonstrates the alignment between source and target.
    }
    \label{fig:al-fail}
\end{figure}
\cref{fig:al-fail} illustrate this phenomenon.
The red line in \cref{fig:al-fail} corresponds to the ideal policy defined in \cite{ma-etal-2019-stacl}. We can see that when the model stops generating the translation, the lagging behind the ideal policy is negative. This is because the model stops reading any input after completing hypothesis generation.
This kind of model can obtain relatively good latency-quality trade-offs as measured by AL (and BLEU), which does not reflect the reality.
We thus define
\begin{equation}
  d^*_i = (i - 1) \cdot \sum_{j=1}^{|\mX|}T_j\ / |\mY^*|
\end{equation}
to prevent this issue, i.e., it is assumed that the ideal policy generates the reference rather than the system hypothesis. The newly defined ideal policy is represented by the green line in \cref{fig:al-fail}.

\item[Differentiable Average Lagging] for the speech task still uses \cref{eq:dal} and \cref{eq:dal-2} with a new $\gamma$ defined as
\begin{equation}
    \gamma_{\text{speech}} = |\mY| / \sum_{j=1}^{|\mX|}T_j
    \label{eq:dal_speech}
\end{equation}
\end{description}
\begin{figure*}[t!]
   \centering
   \includegraphics[width=0.8\textwidth]{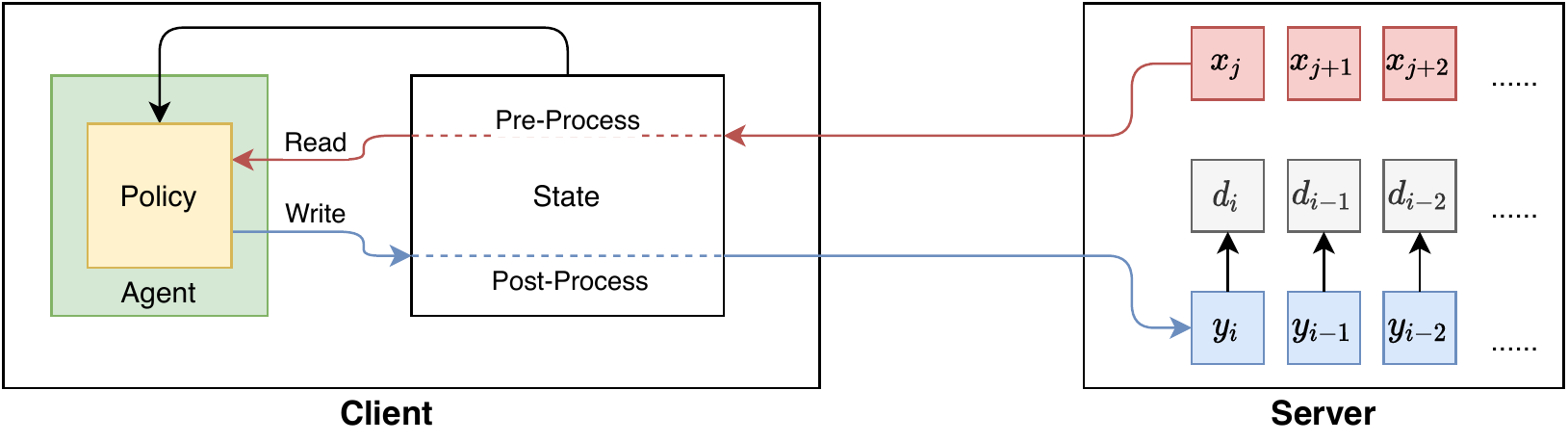}
   \caption{The architecture of \simulevalnsp. The client executes the policy and the server operates the evaluation.}
   \label{fig:architecture}
\end{figure*}
\section{Architecture}
\simuleval simulates a real-time scenario by setting up a server and a client.
The server and client can be run separately or jointly, and are connected through RESTful APIs.
An overview is shown in \cref{fig:architecture}.

\subsection{Server}
The server has primarily four functions.
First, read source and reference files.
Second, send source segments to the client upon a {\tt READ} action.
Third, receive predicted segments from the client upon a {\tt WRITE} action, and record the corresponding delays.
Fourth, run the evaluation on instances. 

The evaluation process by the server on one instance is shown in \cref{alg:server}.
Note that in line 18 in \cref{alg:server}, the server only runs sentence-level metrics. 
The server will collect $\mY$, $\mD$ and $\mT$ for every instance in the evaluation corpus,
and calculate corpus-level metrics after all hypotheses are complete.
\begin{algorithm}
    \small
    \caption{Server side algorithm}
    \label{alg:server}
     \hspace*{\algorithmicindent} \textbf{Input}: $\mX = [x_1,...,x_{|\mX|}], \mY^*=[y^*_1,...,y^*_{|\mY^*|}]$ \\
     \hspace*{\algorithmicindent} \textbf{Input}: $\mY=[], \mD=[]$ \\
      \hspace*{\algorithmicindent} \textbf{Input}: $i=0, j=0, y_0 =\text{BOS}, d_0=0$
    \begin{algorithmic}[1]
    \While{$y_i \neq \text{EOS}$}
        \State $r = \text{await\_request\_from\_client()}$
        \If {$r.\text{action} == \text{READ}$}
            \If  {$j < |\mX|$}
                \State $j = j + 1$
                \State send\_segment\_to\_client($x_j$)
            \Else
                \State send\_segment\_to\_client(EOS)
            \EndIf
        \Else
            \State $i = i + 1$
            \State $y_i = r.\text{segment}$
            \State $\mY = \mY + [y_i]$
            \If {data type is speech}
                \State $d_i = d_{i-1} + T_j$
            \Else
                 \State $d_i = j$
            \EndIf
            \State $\mD = \mD + [d_i]$
        \EndIf
    \EndWhile

    \State \textbf{return} $\text{evaluate}(\mY, \mY^*, \mD, \mT)$
    \end{algorithmic}
\end{algorithm}

\subsection{Client}
The client
contains two components --- an agent and a state. The agent is a user-defined class that
operates the policy and generates hypotheses for simultaneous translation, the latter provides functions such as pre-processing, post-processing and memorizing context. The purpose of this design is to make the user free from complicated setups,
and focus on the policy.
The client side algorithm is shown in \cref{alg:client}.
\begin{algorithm}
    \small
    \caption{Client side algorithm}
    \label{alg:client}
    \hspace*{\algorithmicindent} \textbf{Input}: $\mX = [], i=0, j=0, y_0 =\text{BOS}$, State, Agent
    \begin{algorithmic}[1]
    \While{$y_i \neq \text{EOS}$}
        \State $\text{action} = \text{Agent.policy(State)}$
        \If {$\text{action} == \text{READ}$}
        \State $x = \text{request\_segment\_from\_server()}$
            \If  {$x$ is not EOS}
                \State $j = j + 1$
                
                \State $x_j =  \text{State.preprocess}(x)$
                \State States.update\_source($x_j$)
                \State continue
            \EndIf
        \EndIf

        \State $i = i + 1$
        \State $y_i = \text{Agent.predict(State)}$
        \State $y_i = \text{State.postprocess}(y_i)$
        \State States.update\_target($y_i$)
        \State send\_segment\_to\_server($y_i$)
    \EndWhile
    \end{algorithmic}
\end{algorithm}


\section{Usage Instructions}
\subsection{User-Defined Agent}
A user-defined agent class is required for evaluation,
along with the user's model specific arguments.
The user is able to add customized arguments and initialize the model. Two functions must be defined in order to successfully run online decoding.
The first one is ``policy'',
which takes the state as input and returns a decision on whether to perform a \textit{read} or \textit{write} action.
The other function is ``predict'' which will be called when the ``policy'' returns a \textit{write} action
and return a new target prediction given the state.
An example of a text wait-$k$ model is shown below.
\newpage
\begin{code}
\begin{minted}[
    frame=single,
    framesep=2mm,
    baselinestretch=1.0,
    fontsize=\scriptsize,
    breaklines
    %caption={An example of user defined agent class for a text wait-$k$ model.},
    %captionpos=b,
    %label={code:agent}
]{python}
from simuleval.agents import TextAgent
from simuleval import READ_ACTION, WRITE_ACTION, DEFAULT_EOS
# User defined model code
from user_library import init_model

class WaitKTextAgent(TextAgent):
    def __init__(self, args):
        super().__init__(args)
        # Initialization
        self.waitk = args.waitk
        self.model = init_model(args.model)

    @staticmethod
    def add_args(parser):
        # Customized arguments
        parser.add_argument(
            "--waitk", type=int,
            help="Lagging between source and target")
        parser.add_argument(
            "--model", help="model specifics")

    def preprocess(self, state):
        # preprocess code
        return state

    def postprocess(self, state):
        # postprocess code
        return state

    def policy(self, state):
        # Make a decision here
        if (
            len(state.source) - len(state.target)
            < self.waitk
            and not state.finish_read()
        ):
            return READ
        else:
            return WRITE

    def predict(self, state):
        # Predict a token here
        # Called when self.policy() returns WRITE_ACTION
        return model.predict(state)

\end{minted}
\caption{An example of user defined agent class for a text wait-$k$ model.}
\end{code}

Additionally, the user can define pre-processing or post-processing methods to handle
different types of input.
For example, for a speech translation model,
the pre-processing method can be a feature extraction function
that converts speech samples to filterbank features while for text translation, the pre-processing can be tokenization or subword splitting.
Post-processing can implement functions such as merging subwords and detokenization.

\subsection{User-Defined Client}
A typical user will only need to implement an agent and will rely on the out-of-the-box client implementation of \cref{alg:client}.
However, sometimes, a user may want to customize the client, for example if they want to use a different programming language than Python or make the implementation of \cref{alg:client} more efficient. In that case, they can take advantage of the RESTful APIs between the client and the server described in \cref{tab:restful-api}.
Users can easily plug in these APIs into their own client implementations.

\begin{table*}[h]
    \small
    \centering
    \begin{tabular}{l|l|l|l}
       Function & Endpoint & Params & Response / Body \\
       \midrule
       Get next source segment $x_j$ & /src& \{sent\_id: sent\_idx\} & $x_j$ (text) \\
       Get next $T_j$ ms speech segment $x_j$ & /src & \{sent\_id: sent\_idx, segment\_size: $T_j$\}& $x_j$ (samples) \\
       Send a predicted $y_i$ & /hypo& \{sent\_id: sent\_idx\} & $y_i$
    \end{tabular}
    \caption{A subset of the RESTful APIs for the \simuleval server.}
       \label{tab:restful-api}
\end{table*}

\begin{figure*}[h]
    \centering
    \includegraphics[width=0.75\textwidth]{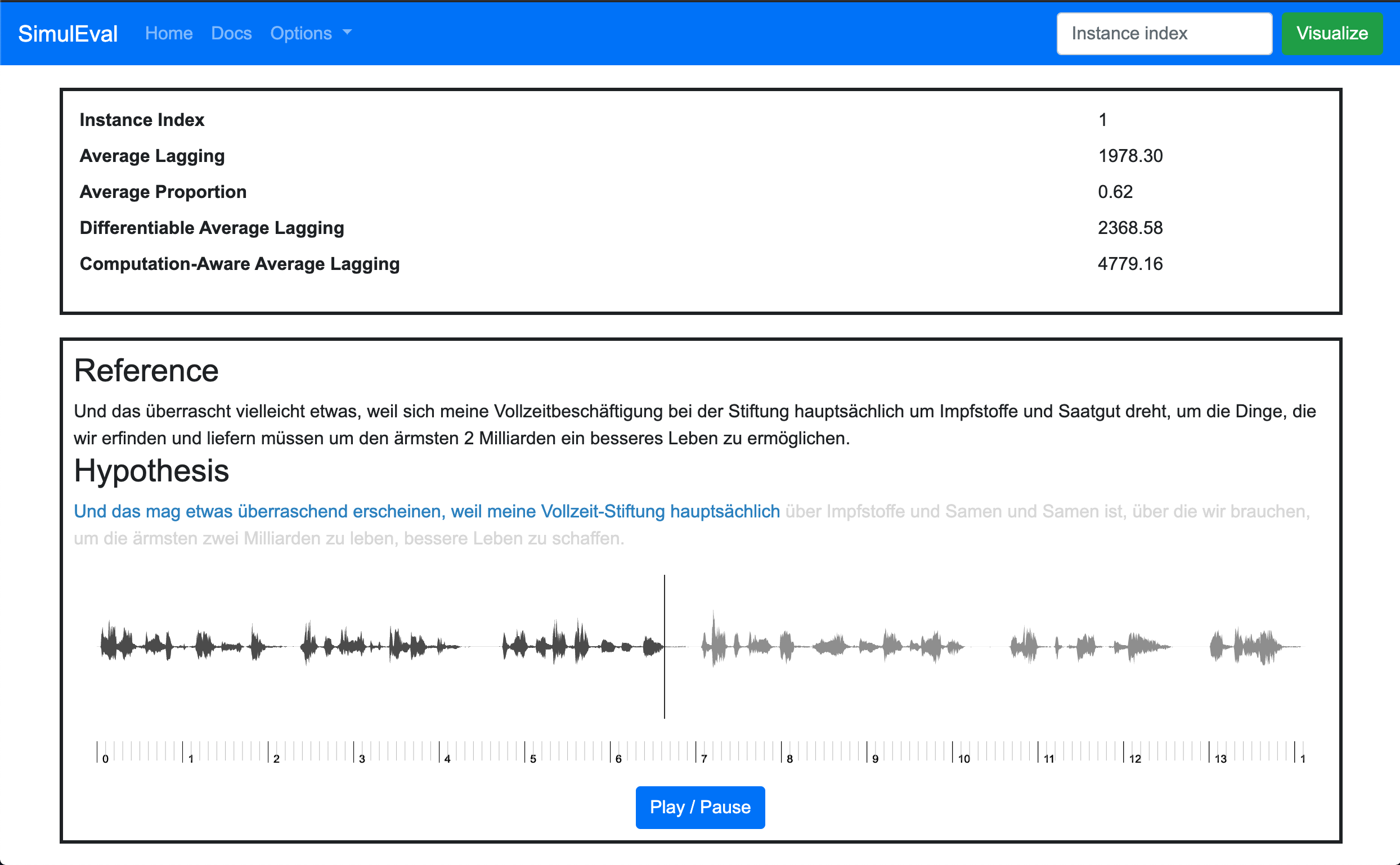}
    \caption{Visualization interface of \simulevalnsp.}
    \label{fig:visual}
\end{figure*}
\subsection{Evaluation}
With a well-defined agent class, \simuleval is able to start the evaluation automatically.
Assuming the agent class is stored in {\tt text\_waitk\_agent.py}, the evaluation can be run in one single command or separate commands:
\begin{code}
\begin{minted}[
    frame=single,
    framesep=2mm,
    baselinestretch=1.0,
    fontsize=\scriptsize,
    breaklines,
    %caption={An example of user defined agent class for a text wait-$k$ model.},
    %captionpos=b,
    %label={code:agent}
]{bash}
simuleval \
  --output $OUT_DIR \
  --source $SOURCE \
  --reference $TARGET \
  --agent text_waitk_agent.py \
  --waitk 3 \
  --model $MODEL_PARAMS
\end{minted}
\caption{Evaluation command (joint)}
\end{code}
\begin{code}
\begin{minted}[
    frame=single,
    framesep=2mm,
    baselinestretch=1.0,
    fontsize=\scriptsize,
    breaklines,
    %caption={An example of user defined agent class for a text wait-$k$ model.},
    %captionpos=b,
    %label={code:agent}
]{bash}
simuleval server \
  --output $OUT_DIR \
  --port 5000 \
  --source $SOURCE \
  --reference $TARGET &

simuleval client \
  --port 5000 \
  --agent text_waitk_agent.py \
  --waitk 3 \
  --model $MODEL_PARAMS
\end{minted}
\caption{Evaluation command (Separate)}
\end{code}

After all hypotheses are generated,
the intermediate results and corpus level evaluation metrics will be saved in the output directory. \simuleval also supports resuming an evaluation if the process has been interrupted.

\subsection{Visualization}
\simuleval provides a web user interface (UI) for visualizing the online decoding process.
\cref{fig:visual} shows an interactive example on simultaneous speech translation.
A user can move the cursor to find the corresponding translation at a certain point.
The visualization server can be simply started by
\begin{minted}[
    frame=single,
    framesep=2mm,
    baselinestretch=1.2,
    fontsize=\scriptsize,
    breaklines,
    %caption={An example of user defined agent class for a text wait-$k$ model.},
    %captionpos=b,
    %label={code:agent}
]{bash}
    simuleval server --visual --log-dir $OUT_DIR
\end{minted}

The default port is 7777 and the web UI can be accessed at \url{http://ip-of-server:7777}.

\subsection{Case Study: IWSLT 2020}
In order to avoid inconsistencies in how latency metrics are computed and to ensure fair comparisons between results presented in research papers, we encourage the research community to use \simuleval when reporting latency in the future.

In addition, an earlier version of \simuleval
was used in the context of the first simultaneous speech translation shared task at IWSLT~\cite{ansari-etal-2020-findings}, where it is of paramount importance to have the same evaluation conditions for all submissions.
In order to preserve the integrity of the evaluation process, the test set, including the source side, could not be released to participants. This motivated the client-server design, where
participants defined their own agent file and submitted their system in a Docker \cite{merkel2014docker} environment. The organizers of the task were then able to run \simuleval and score each submission in a consistent way, even for systems implemented in different frameworks.


\section{Conclusion}
In this paper, we introduced \simulevalnsp,
a general and easy-to-use evaluation toolkit for simultaneous speech and text translation.
It simulates a real-time scenario with a server-client scheme and automatically evaluates
simultaneous translation given a user-defined agent, both for text and speech.
Furthermore, it provides a visualization interface for the user to track the online decoding process.
We introduced example use cases of the toolkit and showed that its general design allows evaluation on different frameworks. 
We encourage future research on simultaneous speech and text translation to make use of this toolkit in order to obtain an accurate and standard comparison of the latency between different systems.
\bibliography{bibliography/emnlp2020, bibliography/simultaneous-translation, bibliography/machine-translation, bibliography/speech-translation}
\bibliographystyle{acl_natbib}
\end{document}